\documentclass{article}

\usepackage{microtype}
\usepackage{graphicx}
\usepackage{subfigure}
\usepackage{booktabs} 
\usepackage{amsthm}
\usepackage{amsmath,amssymb}
\usepackage{color}
\usepackage{bm}
\usepackage{natbib}
\usepackage{cuted}
\usepackage[colorlinks,citecolor=blue,urlcolor=blue,filecolor=blue,backref=page]{hyperref}
\usepackage{graphicx}
\usepackage{mathrsfs}
\usepackage{siunitx}
\usepackage[linesnumbered, algoruled, longend,boxed]{algorithm2e}
\usepackage{caption}
\newcommand{\mmu}{\bm{\mathit{\mu}}}
\newcommand{\mtheta}{\bm{\mathit{\theta}}}

\newcommand{\mXi}{\bm{\xi}}
\newcommand{\mxi}{\mathit{\xi}}

\newcommand{\mSigma}{\bm{\Sigma}}

\newcommand{\mPr}{\Pr}

\newcommand{\mpi}{\mathit{\pi}}

\newcommand{\mz}{\mathbf{z}}

\newcommand{\mD}{\mathbf{D}}

\newcommand{\mS}{\mathit{S}}
\newcommand{\mH}{\mathit{H}}

\newcommand{\mZ}{\mathbf{Z}}

\newcommand{\mR}{\mathbf{R}}

\newcommand{\hmz}{\hat{\mathbf{z}}}
\newcommand{\hmZ}{\hat{\mathbf{Z}}}

\newcommand{\calL}{\mathcal{L}}

\newcommand{\calE}{\mathcal{E}}

\newcommand{\calP}{\mathcal{P}}
\newcommand{\calD}{\mathcal{D}}
\newcommand{\calM}{\mathcal{M}}
\newcommand{\calN}{\mathcal{N}}

\newcommand{\mathR}{\mathbb{R}}

\usepackage[accepted]{icml2019}
\icmltitlerunning{Bayesian surrogate learning in dynamic simulator-based regression problems}

\begin{document}

\twocolumn[
\icmltitle{Bayesian surrogate learning in dynamic simulator-based regression problems}




\begin{icmlauthorlist}
\icmlauthor{Xi Chen}{cam}
\icmlauthor{Mike Hobson}{cam}

\end{icmlauthorlist}

\icmlaffiliation{cam}{Cavendish Laboratory, University of Cambridge, Cambridge, UK}

\icmlcorrespondingauthor{Xi Chen}{xc253@cam.ac.uk}
% \icmlcorrespondingauthor{Mike Hobson}{mph@mrao.cam.ac.uk}

\icmlkeywords{Bayesian, simulator-based system, nested sampling, parameter estimation, deep recurrent networks}

\vskip 0.3in
]



\printAffiliationsAndNotice{}  % leave blank if no need to mention equal contribution

\begin{abstract}
The estimation of unknown values of parameters (or hidden variables, control variables) that characterise a physical system often relies on the comparison of measured data with synthetic data produced by some numerical simulator of the system as the parameter values are varied. This process often encounters two major difficulties: the generation of synthetic data for each considered set of parameter values can be computationally expensive if the system model is complicated; and the exploration of the parameter space can be inefficient and/or incomplete, a typical example being when the exploration becomes trapped in a local optimum of the objection function that characterises the mismatch between the measured and synthetic data.  A method to address both these issues is presented, whereby: a surrogate model (or proxy), which emulates the computationally expensive system simulator, is constructed using deep recurrent networks (DRN); and a nested sampling (NS) algorithm is employed to perform efficient and robust exploration of the parameter space. The analysis is performed in a Bayesian context, in which the samples characterise the full joint posterior distribution of the parameters, from which parameter estimates and uncertainties are easily derived.  The proposed approach is compared with conventional methods in some numerical examples, for which the results demonstrate that one can accelerate the parameter estimation process by at least an order of magnitude.
\end{abstract}

\section{Introduction}
\label{Intro}

Complicated dynamical systems are often modelled using computer-based simulators, which typically depend on a number of hidden or control variables $\mtheta$ that can have a considerable influence on the simulator outputs. The true values $\mtheta_{\ast}$ of these hidden variables are often unknown {\em a priori}, but are usually of considerable interest. These values must therefore be estimated, typically by comparing measured data (or observations) $\calD_{\rm o}$ from the real physical system with deterministic (noise-free) synthetic data $\calD_{\rm s}(\mtheta)$ produced by the simulator as the values of the control variables are varied. Nowadays, this approach is widely used across various sectors including geophysical history matching for reservoir modelling \citep{das2017fast}, earth system modelling for climate forecasting \citep{palmer2012towards}, and cancer modelling and simulation for medical diagnostics \citep{preziosi2003cancer}.

The approach does, however, suffer from two main drawbacks. First, the generation of the synthetic data by the simulator is often computationally intensive. For instance, an oil/gas reservoir simulator can require several days using High Performance Computing (HPC) resources to produce synthetic data for just a single set of parameter values $\mtheta$ \citep{casciano2015latest}. This clearly makes any exploration of the parameter space very slow, particularly for spaces of moderate to high dimensionality, for which one may require hundreds or thousands of simulator runs. Second, the method used to navigate the space of parameters $\mtheta$ may be inefficient and/or lead to incomplete exploration of the space. The parameter estimates $\hat{\mtheta}$ are usually obtained by considering some objective function $F(\mtheta)$ which often exhibit thin degeneracies and/or multiple modes (optima) in the parameter space. For systems where the simulator depends on more than just a few parameters, some standard iterative optimisation algorithm is usually employed (which may or may not require derivatives of the objective function). Such methods typically trace out some path in the parameter space as they converge to the nearest local optimum, rather than the global optimum, and often have difficulty navigating thin degeneracies. In addition, these method provide only a point estimate $\hat{\mtheta}$ of the parameter values, without any quantification of their uncertainty. 

We now outline briefly how the two main drawbacks described above may be circumvented, by the use of model surrogates in combination with a Bayesian approach in which samples are obtained from the posterior distribution of the parameters. We also discuss how such an approach can take advantage of modern parallel computing architectures.

\subsection{Surrogate models}
A system simulator with high computational cost often results in the fact that the process of estimating the hidden parameters, which requires many such evaluations, becomes infeasible in terms of the total runtime required. One potential approach to addressing this issue is simply to exploit large-scale parallel computing resources. Indeed, the rapid progress of GPU hardware development in recent years has made this an attractive prospect. Nonetheless, there still remains the unresolved issues of synchronisation and communication time between CPU/GPU cores, which becomes non-trivial when the number of cores is large. Moreover, parallel computing can only reduce runtime for repeated parallelable sub-tasks, but cannot accelerate individual sub-tasks, or sequential tasks. In particular, dynamical simulators cannot take full advantage of parallel computing because their inherently sequential execution behaviour.

We therefore address the issue of computationally intensive
simulators by constructing a more efficient surrogate model (or
emulator, proxy, meta-model, etc.), which mimics the behaviour of the simulator as closely as possible. This approach is also known as black-box modelling or behavioural modelling since the inner mechanism of the simulator is not assumed to be known. We will focus particularly on surrogate modelling for simulators of dynamical physical systems in regression problem, which typically requires the control variables $\mtheta$ as input only at its initial time step, and then evolves to generate sequential outputs at subsequent time steps. The use of surrogates has the potential to reduce runtime by several orders of magnitude, and naturally accommodates sequential tasks. The main issue with this approach is that the accuracy of the emulator, which depends crucially on the training data and training algorithm. In particular, emulating a very complex simulator often requires a large number of training samples generated from the simulator. Thus, although the runtime for the trained emulator is short, this must be offset against the potentially long time required to prepare the training data and perform the training. Nonetheless, one can mitigate these training costs by taking advantages of parallel computing in some sub-tasks within both training processes.

Over the past two decades, a number of methods have been proposed for simulator-based surrogate construction. These include Gaussian processes (GP, also known as Kriging) \citep{conti2009gaussian, conti2010bayesian, hung2015analysis, mohammadi2018emulating}, neural networks (NN) \citep{van2007fast,melo2014development, tripathy2018deep,  schmitz2018real}, regression based methods and radial basis function (RBF) methods \citep{chen2006review, forrester2009recent}. Of these methods, GP related approaches gain a growing popularity in recent years, but could be computationally infeasible for problems with high dimensionality or with large training data size. NN related methods have shown a good potential, most of its published work, however, targeted non-dynamical problems. Indeed, relatively little work has been performed in applying NN to the construction of surrogates for dynamical simulator-based regression problem. Nonetheless, the recurrent NN (RNN) \citep{goodfellow2016deep} approach has been explored in this context \citep{van2007fast, schmitz2018real}. In particular, \cite{van2007fast}, integrated PCA into the approach to reduce dimensionality across time steps and an RNN was used to predict states in PCA derived space.

Although RNN has proven powerful in many sequence modelling problems (e.g., speech recognition and natural language processing tasks), we have found that it performs poorly in (more general) multivariate dynamical regression problems. This is because such problems often contain a considerable number of hidden (or latent) variables, temporal correlated outputs, and the relationship between the variables and outputs is highly non-linear functions. It is then difficult to generate a corpus of training data that is sufficiently descriptive of the variation of the outputs as a function of the hidden variables. In this paper, we therefore adopt the alternative approach of designing a deep recurrent networks (DRN) under the Bayesian inverse problem framework to address the above issues, the details of which will be discussed in the next section.

\subsection{Bayesian sampler for inverse problem}

Bayesian inference (see e.g. \citealt{mackay2003}) seeks to determine
the posterior probability distribution of a set of unknown parameters
$\mtheta$ in some model $\calM$, given a set of observational data
$\calD_{o}$. This is performed by applying Bayes' theorem:
\begin{align}
\mPr(\mtheta | \calD_{o}, \calM) = \frac{\mPr(\calD_{o} | \mtheta, \calM) \mPr(\mtheta | \calM)}{\mPr(\calD_{o} | \calM)},
\end{align}
where $\mPr(\mtheta | \calD_{o}, \calM) \equiv \calP(\mtheta)$ is the
\texttt{posterior} probability density, $\mPr(\calD | \mtheta, \calM)
\equiv \calL(\mtheta)$ is the \texttt{likelihood} probability density,
$\mPr(\mtheta | \calM) \equiv \mpi(\mtheta)$ is the \texttt{prior}
probability density, and $\mPr(\calD | \calM) \equiv \calE$ is the
\texttt{evidence} (or marginal likelihood). Since $\calE$ is independent of $\mtheta$, one has
\begin{align}
\calP(\mtheta) \propto \calL(\mtheta) \mpi(\mtheta), \label{Sec1:bayeprop}
\end{align}
showing that the posterior is proportional to the product of likelihood and prior. In the context of estimating the parameters $\mtheta$ of a simulator-based model $\calM$ for some physical system, the likelihood provides a measure of the misfit between the observed data and the synthetic data produced by the simulator (or a proxy thereof), as a function of the parameters. This is used to update our prior belief for the parameter values to yield their posterior distribution. In practical scenarios, this typically performed by obtaining samples from the posterior distribution using Monte Carlo numerical methods \citep{robert2004monte}.

\subsubsection{Nested sampling}
\label{Sec:NS}
This paper adopts Nested sampling (NS) \citep{skilling2006nested} as our primary Bayesian sampler, it is a sequential sampling approach that resolves many of the problems encountered by Markov Chain Monte Carlo (MCMC) by evolving a fixed number of points in the space to explore the posterior distribution in a different way. In addition, NS simultaneously provides posterior samples and an estimate of the evidence. Nested sampling algorithm is briefly described in pseudo-code given below.

\begin{algorithm}[!ht]
\tcp{Initialisation}

At iteration $i=0$, draw $ N_{\rm live}$ samples $\{\mtheta_n\}_{n=1}^{N_{\rm live}}$ from prior $\pi(\mtheta)$ within prior space $\Psi$. 

Initialise evidence $Z=0$ and prior volume $X_0 = 1$. 

\tcp{Sampling iterations}

\For{$i=1, 2, \cdots, I$}{
	$\bullet$ Compute likelihood $\calL(\mtheta_{n})$ for all
  $N_{\rm live}$ samples. 
  
	$\bullet$ Find lowest likelihood in live sample and save 
as $\calL_i$. 

	$\bullet$ Calculate weight 
$w_i = \frac{1}{2}(X_{i-1} - X_{i+1})$, where the prior volume $X_{i} = \exp(-i/N_{\rm live})$. 

	$\bullet$ Increment evidence $Z$ by $\calL_i w_i$. 
	
	$\bullet$ Replace the individual sample with likelihood $\calL_i$ by a newly drawn sample from restricted prior space $\Psi_i$ such that $\mtheta \in \Psi_i$ satisfies $\calL(\mtheta) > \calL_i$.
	
    $\bullet$ If $\mbox{max}\{\calL(\mtheta_n)\}X_i < \exp({\tt{tol}})Z$,then \bf{stop}.}

Increment $Z$ by $\sum_{n=1}^{N_{\rm live}}\calL(\mtheta_n)X_{I}  / N_{\rm live}$. 

Assign the sample replaced at iteration $i$ the importance weight $p_i=L_iw_i/Z$.

\caption{Nested sampling algorithm}
\label{alg:NS}
\end{algorithm}

$\Psi$ denotes parameter space of $\mtheta$. $I$ represents total number of iterations, and its value depends on both the pre-defined convergence criteria and the complexity of the problem. $X_0$ is the prior volume at iteration $0$, and $X_i$ denotes the constrained prior volume at $i$th iteration.  

Some widely used NS variants, such as MultiNest \citep{feroz2009multinest,feroz2013importance} and PolyChord \citep{handley2015polychord} have been developed in recent years. In MultiNest, new samples are drawn by rejection sampling at each iteration, from within a multi-ellipsoid bound to an iso-likelihood surface; the bound is computed by samples present at each iteration. It is worth noting that these NS methods can make use of parallel computing resources within the sampling algorithm.

\subsection{The proposed approach}

We develop a deep recurrent network (DRN) technique to train surrogate models for multivariate dynamical simulator-based system, which is then used to perform a fast calculation of synthetic data $\calD_{\rm s}(\mtheta)$ from a given set of control parameters $\mtheta$. $\calD_{\rm s}(\mtheta)$ are then used alongside measured data $\calD_{\rm o}$ from the real physical to provide a rapid evaluation of the likelihood function within a Bayesian analysis. Using this fast likelihood function, the parameter space $\mtheta$ is explored using NS (via the MultiNest algorithm) to obtain samples from the posterior distribution and evidences. 

The DRN employs a temporal cascading structure that is used to model temporal dynamics in simulator. This structure divides the system outputs into temporal features, and constructs a series of cascaded components to form a surrogate model. The proposed approach greatly facilitates the training process, and leads to a runtime at least an order of magnitude smaller than that of the simulator. Moreover, unlike traditional proxy building methods that use random samples for training, in our approach the surrogate model is trained using samples generated by the nested sampling algorithm. This assists in training by providing higher sample density in the high-likelihood region of interests, and thus improves the overall accuracy with which the parameters can be estimated. 

The paper is organised as follows. Section~\ref{Sec:simulator}
introduces and formulates the problems. Section \ref{Sec:cascading} details the design of DRN. Section \ref{Sec:wholeAlg} depicts a complete pipeline of the proposed Bayesian surrogate modelling approach. Section \ref{Sec:NumEval} presents numerical results. Section \ref{Sec:Con} concludes the work and discusses some future directions.

\section{Problem formulation} 
\label{Sec:simulator}
\begin{figure}[ht]
\centering
\includegraphics[width = 0.96\linewidth]{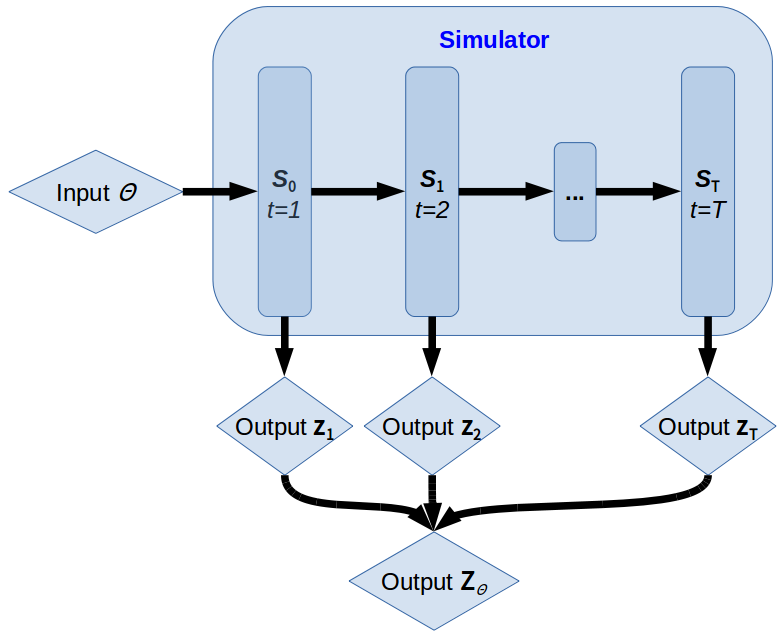}
\caption{Sequential execution in a simulator. The rectangle blocks $\{\mS_{1}, \mS_{2},\cdots,\mS_{T}\}$ represent simulator components at times $t$.}
\label{fig:seqSimulator}
\end{figure}

Consider a simulator $\mS$ with $J$ input and $M$ output features. The simulator only takes inputs $\mtheta$ at time $t = 0$, and evolves for $T$ steps. The total output number is $T \times M$, and the noise-free simulator execution process can be written as:
\begin{align}
\mZ_{\mtheta} = \mS(\mtheta),
\label{Sec2:noisefreeEq}
\end{align}
where input vector $\mtheta \in \mathR^{J \times 1}$, with
$\mtheta = [\theta_1, \cdots, \theta_j, \cdots, \theta_J]^{\intercal}$, and $\mZ_{\mtheta} \in \mathR^{T \times
  M}$  is the simulator output
matrix corresponding to $\mtheta$, with $\mZ_{\mtheta} = [\mz_1, \cdots, \mz_t, \cdots, \mz_T]^{\intercal}$. The column vector $\mz_t$ contains $M$ output features at time $t$, and $\mz_t = [z_{t,1}, \cdots, z_{t,m}, \cdots, z_{t,M}]^{\intercal}$. The  sequential execution process is shown in Figure \ref{fig:seqSimulator}. $\mS_t$ denotes an inner component of $\mS$ producing $\mz_{t}$ at time $t$.

In practice, the simulator output $\mZ_{\mtheta}$ may not correspond to noise-free synthetic data. We denote the synthetic data by the matrix $\mD_{\mtheta}$ (previously $\calD_{\rm s}(\mtheta)$) with the same dimensions and structure as $\mZ_{\mtheta}$, then the process is expressed as $\mD_{\mtheta} = \mR(\mZ_{\mtheta})$. $\mR$ describes the response of the measurement system used, which may be a complicated non-linear function of its argument. In this paper we will assume, however, that the response function of the measurement process is simply the identity, so that $\mD_{\mtheta}$ is equal to $\mZ_{\mtheta}$.

The complete inference of a set of unknown parameters $\mtheta$ from an observed data set $\mD_{o}$ is contained in the posterior
distribution $\mPr(\mtheta | \mD_{o})$, which can be described by a set of discrete samples $\{\mtheta^{(i)}\}_{i=1}^{I}$ with their weights (posterior probability) computed by Equation
\eqref{Sec1:bayeprop}:
\begin{align}
\mPr(\mtheta | \mD_{o}) \sim \{\mtheta^{(i)} ; \mPr(\mD_{o} | \mtheta^{(i)}) \mPr(\mtheta^{(i)}) \}_{i=1}^{I}.
\label{Eq:postSet}
\end{align}
The prior is assumed as a Gaussian with $\mtheta \sim \calN(\mmu_{\mtheta}, \mSigma_{\mtheta})$. We also assume a Gaussian additive noise (non-Gaussian noise model case will be presented in an extended version paper) $\mXi$ to describe non-informative part of the observation, such that $\mxi_{t,m} \sim \calN(0,(\sigma_{t,m}^{\mxi})^2)$ for the $m$th feature at time $t$. Sample likelihood $\mPr(\mD_{o} | \mtheta^{(i)})$ is given by:
\begin{equation}
\calL(\mtheta^{(i)}) = \prod_{t=1}^{T} \prod_{m=1}^{M} 
\left\{ \frac{1}{\sqrt{2 \pi (\sigma_{t,m}^{\mxi})^2}} \exp \left[-\frac{(d_{t,o} - z_{t,m}^{(i)})^2}{2 (\sigma_{t,m}^{\mxi})^2}\right] \right\},
\label{Eq:likelihood}
\end{equation}
where $\mD_{o} = [d_{1,o}, \cdots, d_{t,o}, \cdots, d_{T,o}]$ is the observed data, and $z_{t,m}^{(i)}$ is the simulator prediction for sample $\mtheta^{(i)}$. The posterior $\mPr(\mtheta | \mD_{o})$ can then be calculated by Equation \eqref{Eq:postSet}. 

\section{DRN surrogate construction}
\label{Sec:cascading}

Surrogate model $\mH$ is used to emulate the simulator described in Equation \eqref{Sec2:noisefreeEq}, and is written as:
\begin{align}
\hmZ_{\mtheta} = \mH(\mtheta), \label{Eq:surrogateM}
\end{align}
where $\hmZ_{\mtheta}$ is the surrogate output. The goal is to construct an $\mH$ that minimises the misfit between $\hmZ_{\mtheta}^{(i)}$ and $\mZ_{\mtheta}^{(i)}$ given training set $\{\mtheta^{(i)}, \mZ_{\mtheta}^{(i)}\}_{i=1}^{I}$,. The probability of a single training pair $\mPr(\mZ_{\mtheta}^{(i)}|\mtheta^{(i)})$ can be decomposed as:
\begin{align}
\log \mPr(\mZ_{\mtheta}^{(i)}|\mtheta^{(i)}) = \sum_{t}\log(\mz_t^{(i)}|\mz_{t-1}^{(i)},\mtheta^{(i)}),
\end{align}
where $\mz_t^{(i)}$ is a sequence element in $\mZ_{\mtheta}^{(i)}$ with
\begin{align}
\mz_t^{(i)} = \left\{
                \begin{array}{ll}
                  \mH_{t}(\mtheta^{(i)}), & \text{if } t=1, \\
                  \mH_{t}(\mtheta^{(i)},\mz_{t-1}^{(i)}), & \text{otherwise.}
                \end{array}
              \right.
\label{Eq:cascadeEle}
\end{align}
$\mH$ is also composed by a set of functional sequence components $\{ \mH_{t}\}_{t=1}^{T}$. We define model hyper-parameters as $\Phi$, thus training a surrogate is equivalent to maximising the log-likelihood w.r.t. hyper-parameter set $\Phi$:
\begin{align}
\Phi^{\ast} = \arg \max\limits_{\Phi} \sum_{i = 1}^{I} \log \mPr(\mZ_{\mtheta}^{(i)} | \mtheta^{(i)}; \Phi), 
\label{Eq:maxLogLike}
\end{align}
where $\Phi^{\ast}$ denotes the optimal hyper-parameter set. 

\subsection{Design of DRN}

\begin{figure}[ht]
\centering
\includegraphics[width = 0.8\linewidth]{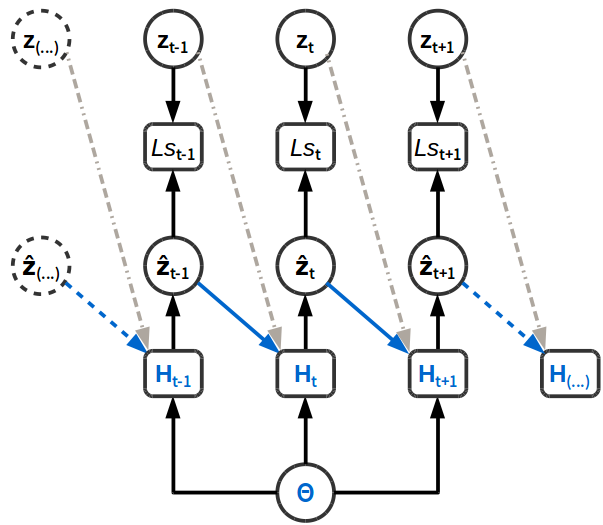}
\caption{A DRN illustrated in unfolded computational
  graph. Rectangle block represents function and round block
  represents variable. Model output is denoted by $\{ \hmz_{t} \}_{t=1}^T$. Loss function in DRN at time $t$ is $Ls_t$, which computes the error between $\hmZ_{\mtheta}$ and the training sample $\mZ_{\mtheta}$. Features of the proposed DRN are highlighted in blue dashed line: (1) the input $\mtheta$ is fixed in our problem. (2) the hidden component $\mH_{t}$ contains an extended DNN structure in DRN. (3) predicted sample $\hmz_{t}$ (solid blue line), rather than training sample $\mz_t$ (dash-doted grey line) is used to feed its next hidden component $\mH_{t+1}$.}
\label{fig:compGraph}
\end{figure}

As illustrated in Figure \ref{fig:compGraph}, the proposed DRN is one of the typical designs \citep{goodfellow2016deep} whose only recurrence is from the prediction $\hmz_t$ to its next hidden component $\mH_{t+1}$. $\mH_t$ in Figure \ref{fig:compGraph} is extended as an independent DNN (conditioned on its previous prediction) to learn features at each time step. A pre-configured DNN (detailed in the next section) is adopted as a component-wise ML training algorithm within the proposed DRN structure. This paper mainly focuses on introducing a complete Bayesian surrogate modelling approach, topics such as DNN structure design, hyper-parameter tuning, DRN design with LSTM, or scheduled sampling \citep{bengio2015scheduled} is beyond the scope of this paper. 

\subsection{DNN hidden component}
\label{Supp:DNN}
\begin{figure}[ht]
\centering
\includegraphics[width = 0.6\linewidth]{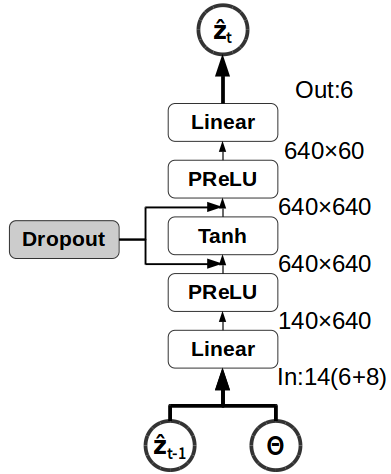}
\caption{A example to illustrate component-wise DNN
  structure. The model complexity factor $\eta$ is set to 10 in this example. Component $\mH_{t}$ has 14 inputs and 6 outputs, and contains $4$ forward hidden layers and $2$ Dropout layers. The
  bottom two circle blocks depict input to the DNN, including $\mtheta$ (with 8 input elements) and $\hmz_{t-1}$ (with 6 predictions). The rectangle blocks denote activation functions
  between different layers. The numbers on the right of the rectangle blocks represent node number of different DNN layer. }
\label{fig:DNN}
\end{figure}

As shown in Figure \ref{fig:DNN}, we employ a DNN structure with 4 hidden layers and 2 Dropout layers. $\mH_t^{(b)}$ denotes $b$th hidden layer, and the number of nodes in each hidden layer depends on the number of inputs $J$, the number of features $M$, and complexity factor $\eta$. Particularly, node number in each DNN layer is determined as follows:
\begin{itemize}
\item Input layer: $J + M$.
\item $\mH_t^{(1)}$: $\eta \times (J+M)$.
\item $\mH_t^{(2)}$: $4 \times \eta \times (J+M)$.
\item $\mH_t^{(3)}$: $4 \times \eta \times (J+M)$.
\item $\mH_t^{(4)}$: $\eta \times M$.
\item Output layer: $M$.
\end{itemize}

Activation function used in the proposed DNN includes $2$ linear functions, $2$ Parametric Rectified Linear Unit (PReLU) functions, and $1$ Hyperbolic tangent (Tanh) function. In the numerical examples, the proposed DNN structure and the majority of its hyper-parameters are fixed. Node number in the hidden layers can be adjusted through the complexity factor $\eta$. We adopt standard mean square error (MSE) as the loss function in DRN training, and use standard \textit{Adam} algorithm (which is a first-order gradient-based stochastic optimisation method) as the optimiser. The Dropout ratio is set to 0.5 for the Dropout layers. 

Although the proposed DRN structure can be generalised to
different problems, constructing a good DNN architecture still highly depends on the complexity and dimensionality of the target problem and simulator. DNN structure needs to be carefully designed with specific purposes by examining a series of properties such as loss function, activation function, optimiser, layer number, node number, and evaluation metrics, etc.. The structure illustrated in Figure \ref{fig:DNN} is one of our design choices that fits well in the numerical examples.

\section{Bayesian surrogate learning}
\label{Sec:wholeAlg}

The pipeline of a complete Bayesian surrogate learning process includes three phases: (1) sampling from simulator; (2) surrogate model construction; and (3) sampling from surrogate model. Details are described as follows and also illustrated in supplementary material Section Figure \ref{fig:BayeProc}.

\subsection{Sampling from simulator}
The goal in Phase 1 is to generate posterior samples for surrogate training through Bayesian sampling techniques performed in the original computationally intensive simulator $\mS$. MultiNest firstly draws $\{ \mtheta^{(i)}\}_{i=1}^{I}$ random samples from prior $\mpi(\mtheta)$. Simulator outputs $\mZ_{\mtheta}$ and its corresponding likelihood $\calL(\mtheta)$ can then be computed through Equation \eqref{Sec2:noisefreeEq} and \eqref{Eq:likelihood}, respectively. One can obtain posterior samples $\{\mtheta^{(i)}\}_{i=1}^I$ and their weights $\mPr(\mtheta|\mD_{o})$ of $\mtheta$ by Equation \eqref{Eq:postSet}.

In practice, it is often time consuming to use posterior samples for training. Collecting Bayesian posterior samples requires sampling algorithms to perform a considerable (but unknown) number of likelihood evaluations. An alternative way is to simply draw random samples from the prior. Prior can be in uniform, Gaussian, or Latin-Hyper-Cube (LHC) \citep{mckay1979comparison} etc, and this method only requires exactly $I$ likelihood evaluations. One key drawback of random prior drawn, however, relies on its flat sample densities in both the high and low likelihood regions, which is clearly less efficient than the posterior drawn approach when exploring high likelihood regions.

\subsection{Surrogate model training}
The obtained posterior samples $\{\mtheta^{(i)}\}_{i=1}^I$ are then fed as input to simulator $\mS$ to generate $\mZ_{\mtheta}$. The training set $\{\mtheta^{(i)},
\mZ_{\mtheta}^{(i)}\}_{i=1}^{I}$ can then be collected for surrogate training for the proposed DRN described in Section
\ref{Sec:cascading}.

Bayesian posterior samples often concentrate in high likelihood regions, in which they can provide high resolution for surrogate learning. However, this also leads to its poor generalisation in the whole state space. On the contrary, random drawn samples can well generalised in the state space,
but often require a relatively bigger sample size to gain enough resolution for the region of interests. This issue is further discussed and illustrated in a numerical example.

\subsection{Sampling from surrogate model}
Once the surrogate model $\mH$ is obtained, it can be used for fast parameter calibration following a similar procedure as described in Phase 1. In fact, the surrogate acts as a fast forward mapping between $\mtheta$ and $\hmZ_{\mtheta}$, and the DRN is used to interpolate the contour space based on the training data. One of the constrains for this approach is in training data collection step. Even with random drawn samples, the approach still requires simulator $\mS$ to execute $I$ times for training data collection. A small $I$ can lead to poor generalisation, while a big $I$ results in a long data preparation time. The choice of $I$ varies case-by-case, and a trade-off should be carefully considered in different practical cases to balance estimation accuracy and total
runtime.

\section{Numerical examples} \label{Sec:NumEval}

This section reports numerical performance of the proposed method. It is implemented with Keras (TensorFlow backend) \citep{chollet2015keras} in Python 3.5 environment. In particular, we compare: (a) the effect of using different training sample sources including Latin Hyper-Cube (LHC) samples \citep{mckay1979comparison}, MultiNest posterior samples, and a mix of them. (b)Performance between the proposed DRN, a standard DNN, and a standard RNN. (c) Total runtime between simulator-based method and surrogate method using a same sampler. In the tests, training process is executed by an NVIDIA Quadro K2200 GPU with 640 CUDA cores and 4 GB graphical RAM. MultiNest estimation is performed in a computer equipped by an Intel Xeon E3-1246 v3 CPU with
4 cores (3.5 GHz) and 16 GB RAM.

\subsection{Bivariate dynamic model}
\label{exp:toy1}

\begin{figure}[ht]
\centering
\includegraphics[width = 1.02\linewidth]{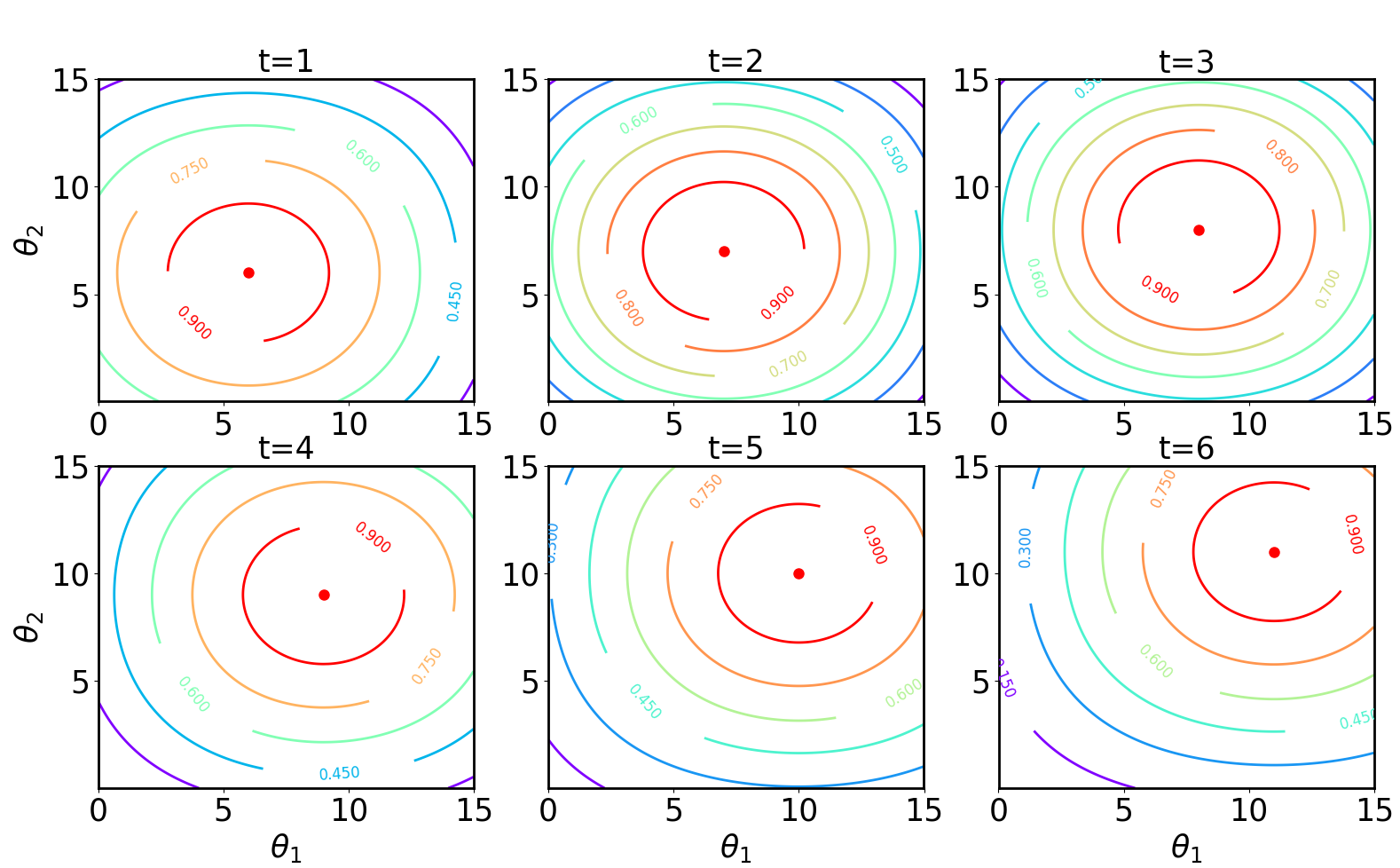}
\caption{Illustration of temporal dynamics of simulator output in the bivariate example.The sub-figures depict contour from time $t=1$ to $6$. The value inline of contour denotes output $\mz_t$ of that contour circle, and the peak is highlighted in red dot at centre.}
\label{fig:Exp1_Contour}
\end{figure}

A synthetic 2 inputs and 10 outputs toy simulator is constructed to benchmark the algorithm performance. It has $J=2$ unknown parameters, and $M=1$ output feature for $T=10$ time steps (thus $10$ outputs in total). The simulator function is defined as:
\begin{align}
\mz_{t} = \cos[\phi (\theta_{1} - t - \eta)] \cos[ \phi (\theta_{2} - t - \eta)],
\end{align}
where input $\mtheta = [\theta_{1}, \theta_{2}]$, $\mz_{t}$ denotes simulator output at time $t$, and
$\mz_{t} = z_{t}$ in this example. $\phi$ and $\eta$ are known constant coefficients, with $\phi = 0.1$, and $\eta = 5$. Prior range is set to $\mtheta \in [0, 15]$, and the ground truth $\mtheta_{\ast} = [\theta_{\ast1}, \theta_{\ast2}]$ are $\theta_{\ast1} = \theta_{\ast2} = 10$. It is a uni-mode toy simulator, of which its highest output (the peak) gradually moves toward north-east direction in the 2D parameter space whereas the contour shape remains the same, as shown in Figure \ref{fig:Exp1_Contour}. 

\begin{figure*}[ht]
\subfigure[Surrogate testing RMSE]{
\includegraphics[width = 0.32\linewidth]{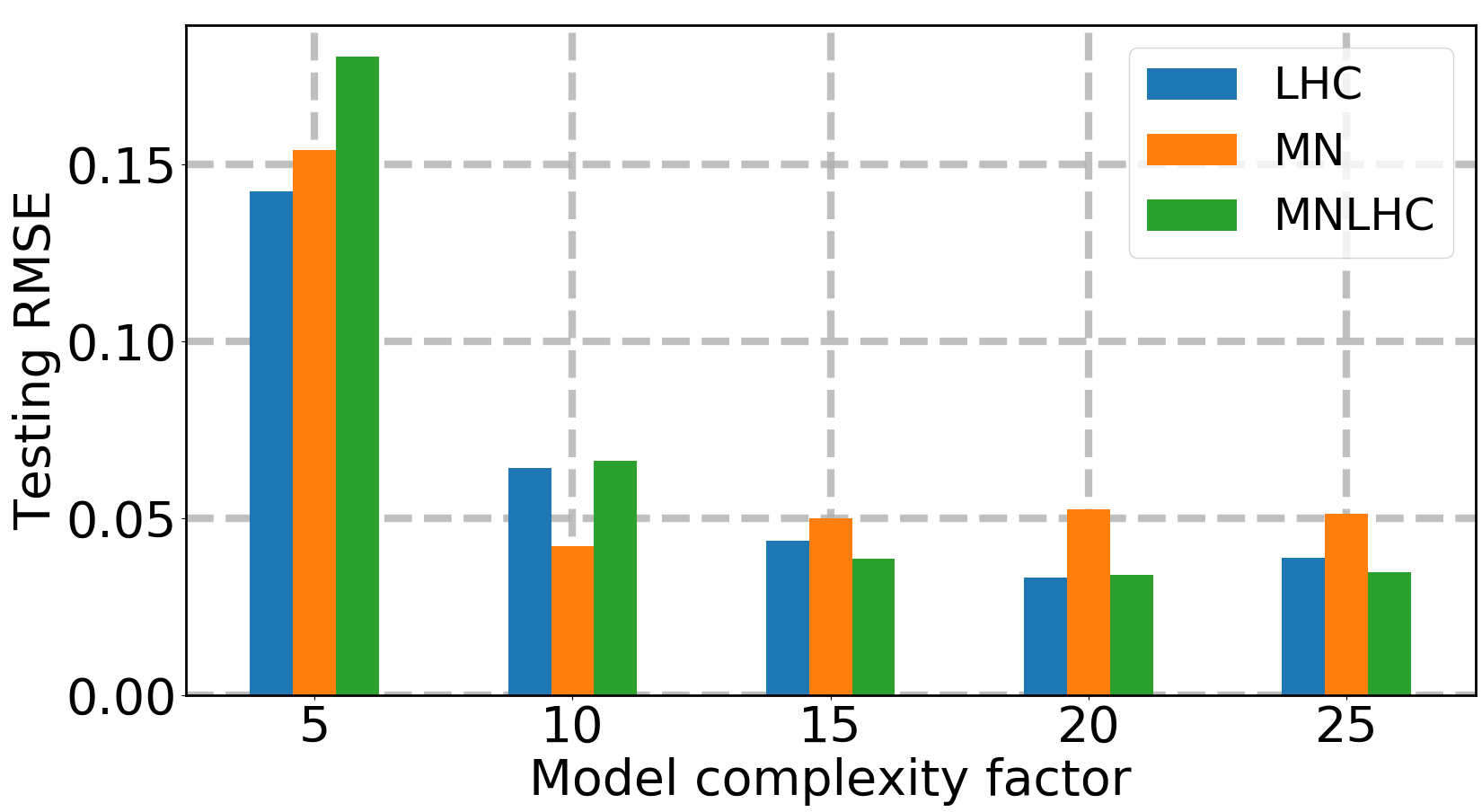}}
\subfigure[Surrogate testing correlation]{
\includegraphics[width = 0.32\linewidth]{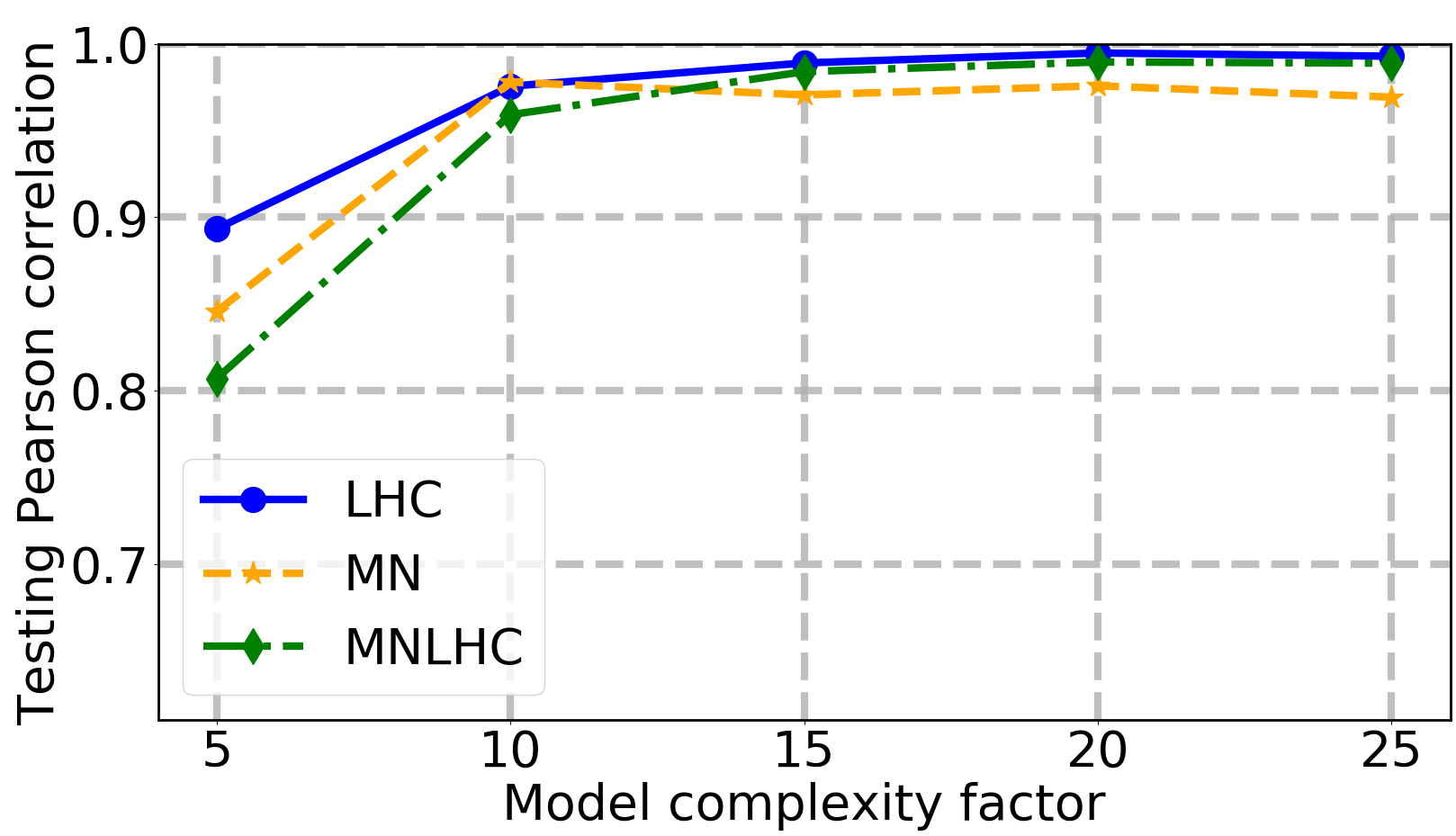}}
\subfigure[Estimation RMSE]{
\includegraphics[width = 0.32\linewidth]{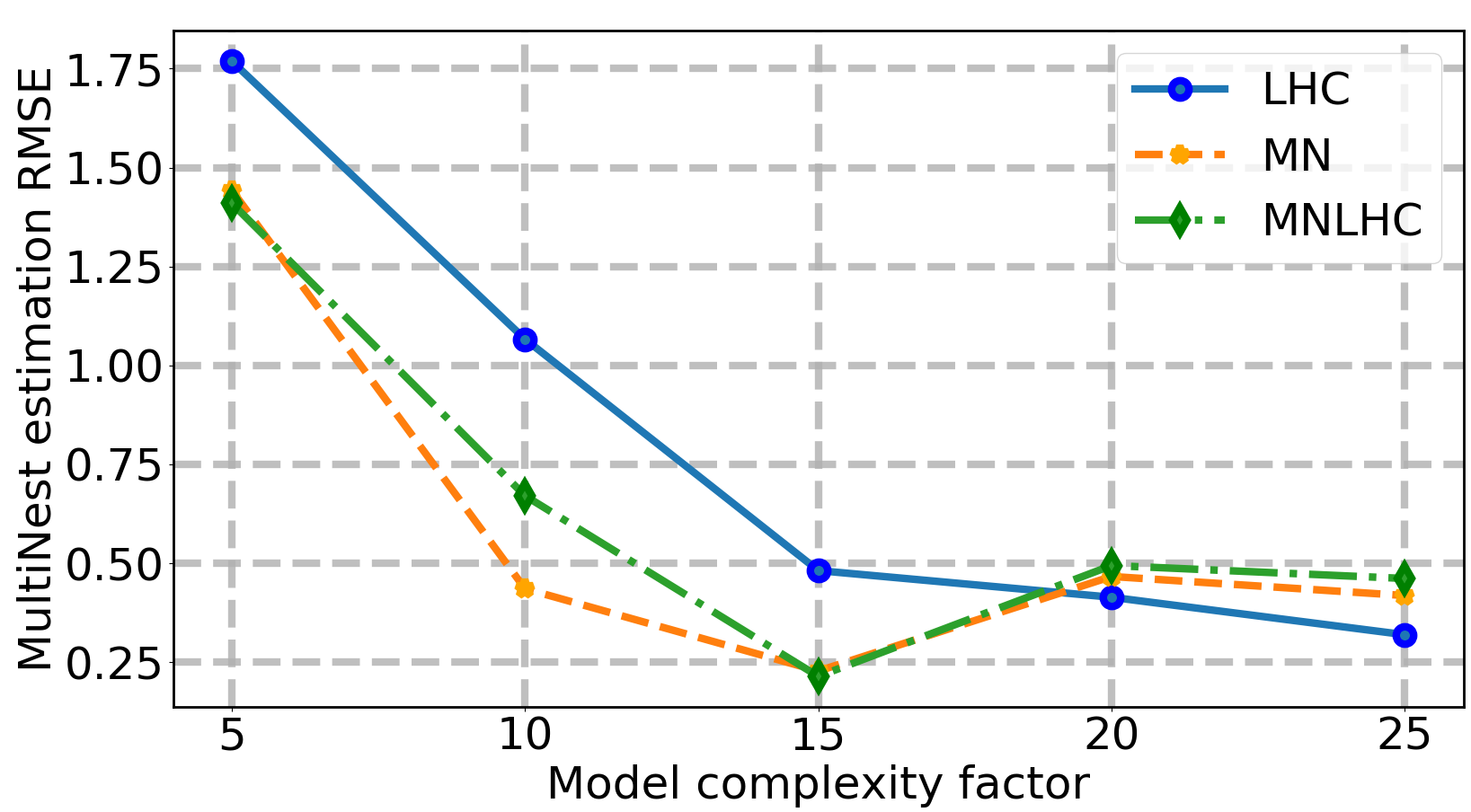}}
\caption{Accuracy comparison of DRN surrogate with respect to model complexity $\eta$ in different training sample sources. A bigger value of $\eta$ denotes a more complex DNN structure in the proposed DRN. LHC: Latin Hyper Cube; MN: MultiNest posterior; MNLHC: an equal mix of MultiNest and LHC samples. Surrogate model accuracy is evaluated in RMSE (sub-figure (a)), and Pearson correlation (sub-figures (b)), both using out of sample testing data set. All results are evaluated with 10-folds cross-validation. Sub-figure (c) shows final estimation accuracy of $\mtheta$ (in RMSE) using corresponding trained surrogate in MultiNest.}
\label{fig:toyRMSE}
\end{figure*}

\subsubsection{Choice of training sample source}

Industrial system simulator often contains a number of non-linear functions, which may result in highly complex contour in parameter space. A suitable training sample collection is then desired in this situation to improve surrogate model accuracy. Here we compare $3$ different training sample sets:
\begin{enumerate}
\item Latin Hyper-Cube (LHC) samples. 
\item Posterior samples from MultiNest algorithm.
\item An equal mix of MultiNest and LHC samples. 
\end{enumerate}

The number of training sample $N$ is set to $2000$, and for case (3) samples are composed by $N/2$ LHC samples and $N/2$ MultiNest posterior samples. Other settings are fixed: the Gaussian noise standard deviation $\sigma_{t,m}^{\mxi}$ is set to $5\%$ of the averaged
noise-free simulator output $\mZ_{\mtheta}$, and the prior of $\mtheta$ is a uniform distribution with range $[0,15]$. DRN is trained with $300$ epochs and its mini-batch size is $20$. A 10-fold cross-validation (CV) method is used for training, i.e., the ratio between training and testing data is $9:1$.

Figure \ref{fig:toyRMSE} (a) and (b) shows accuracy comparison of different trained surrogate in test data. One sees that the surrogate accuracy (represented in root mean squared error (RMSE) and Pearson Correlation) is affected by different training data collection schemes. Both RMSE and correlation are calculated between the true training outputs and the surrogate predicted outputs. Surrogate accuracy stays at a robust level when $\eta \geq 15$. In sub-figure (b) all surrogate models score high correlation, i.e, $> 95\%$ when $\eta \geq 10$. The results demonstrate that the trained models can achieve comparable accuracy for $\eta \geq 15$ for the bivariate example. 

\begin{figure}[ht]
\subfigure[MultiNest posterior]{
\centering
\includegraphics[width = 0.48\linewidth]{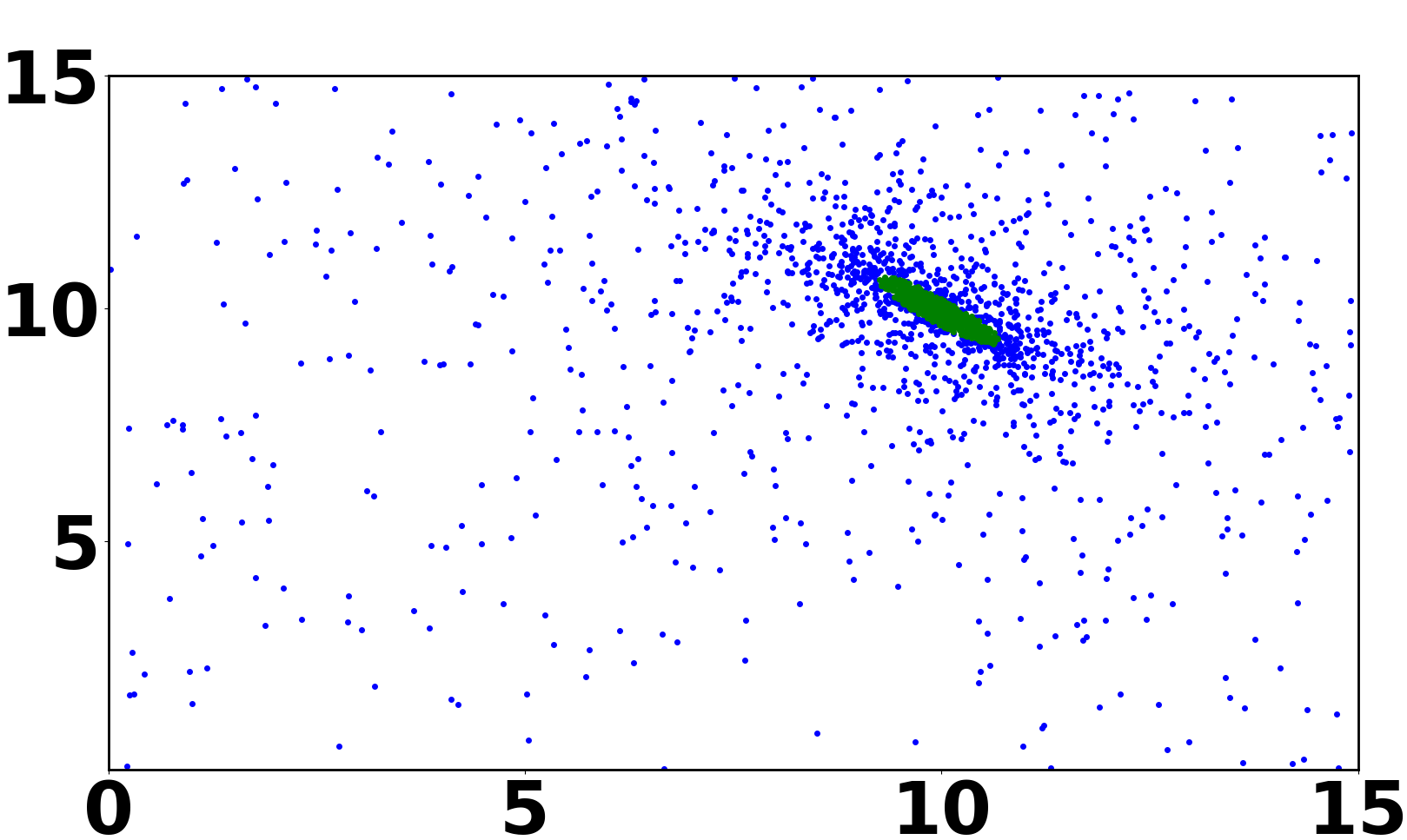}}
\subfigure[MultiNest + LHC]{
\centering
\includegraphics[width = 0.48\linewidth]{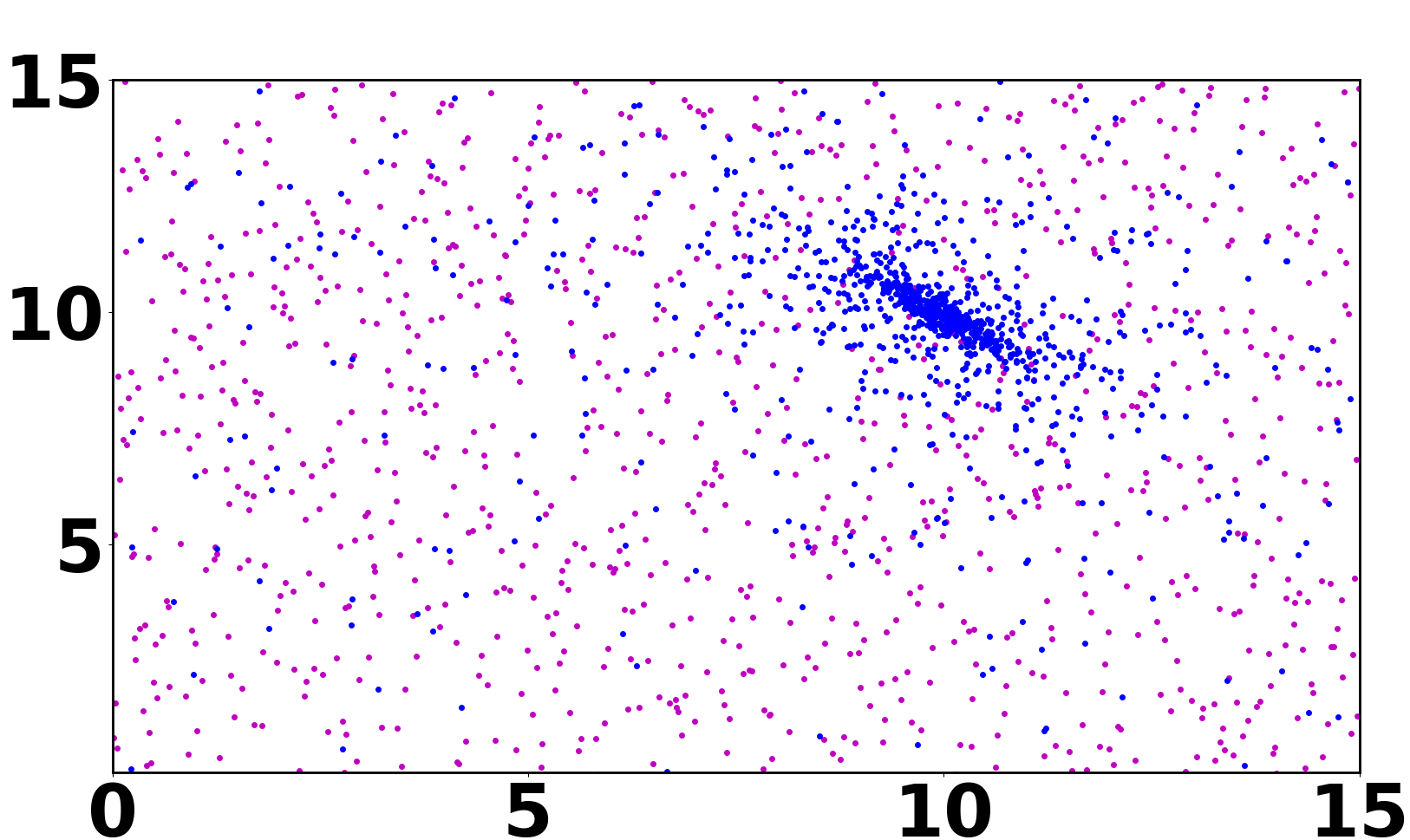}}
\caption{Illustration of MN (sub-figure (a)) and MNLHC samples (sub-figure (b)). MultiNest samples are denoted in green and blue (darker in grey-scale). LHC samples are denoted in purple (lighter in grey scale).}
\label{fig:MCMCSamples}
\end{figure}

Figure \ref{fig:toyRMSE} (c) shows the corresponding MultiNest estimation performance. The best estimation accuracy is achieved by case (2) and (3) samples, both with model complexity $\eta = 15$. This is different from the observation in Figure \ref{fig:toyRMSE} (a), where the best performance is achieved at $\eta = 20$ by case (1) and (3) samples. The observation suggests that highly accurate surrogate doesn't necessarily guarantee best posterior estimation performance. In fact, the two accuracy indicators (namely, surrogate RMSE and estimation RMSE) will give more consistent performance when the sampling resolution in parameter space is high enough. 

The performance shows that training sample bias can affect surrogate model performance. Figure \ref{fig:MCMCSamples} illustrates the difference between MN and LHC sampling schemes. In sub-figure (a), MultiNest initialises random samples (blue dots) across the parameter space, and samples intensively move toward the high likelihood region (green dots). Sub-figure (b) shows a mix of MN and LHC samples (purple dots), where LHC samples help to describe and generalise the whole parameter space for surrogate training. In our empirical tests, an equally mixed MCMC posterior and LHC samples (or uniform samples) also achieves similar results as those in case (3) MNLHC, which suggests that the proposed scheme is suitable for other Bayesian posterior sampling algorithms. 

\begin{figure*}[ht]
\subfigure[RMSE]{
\centering
\includegraphics[width = 0.32\linewidth]{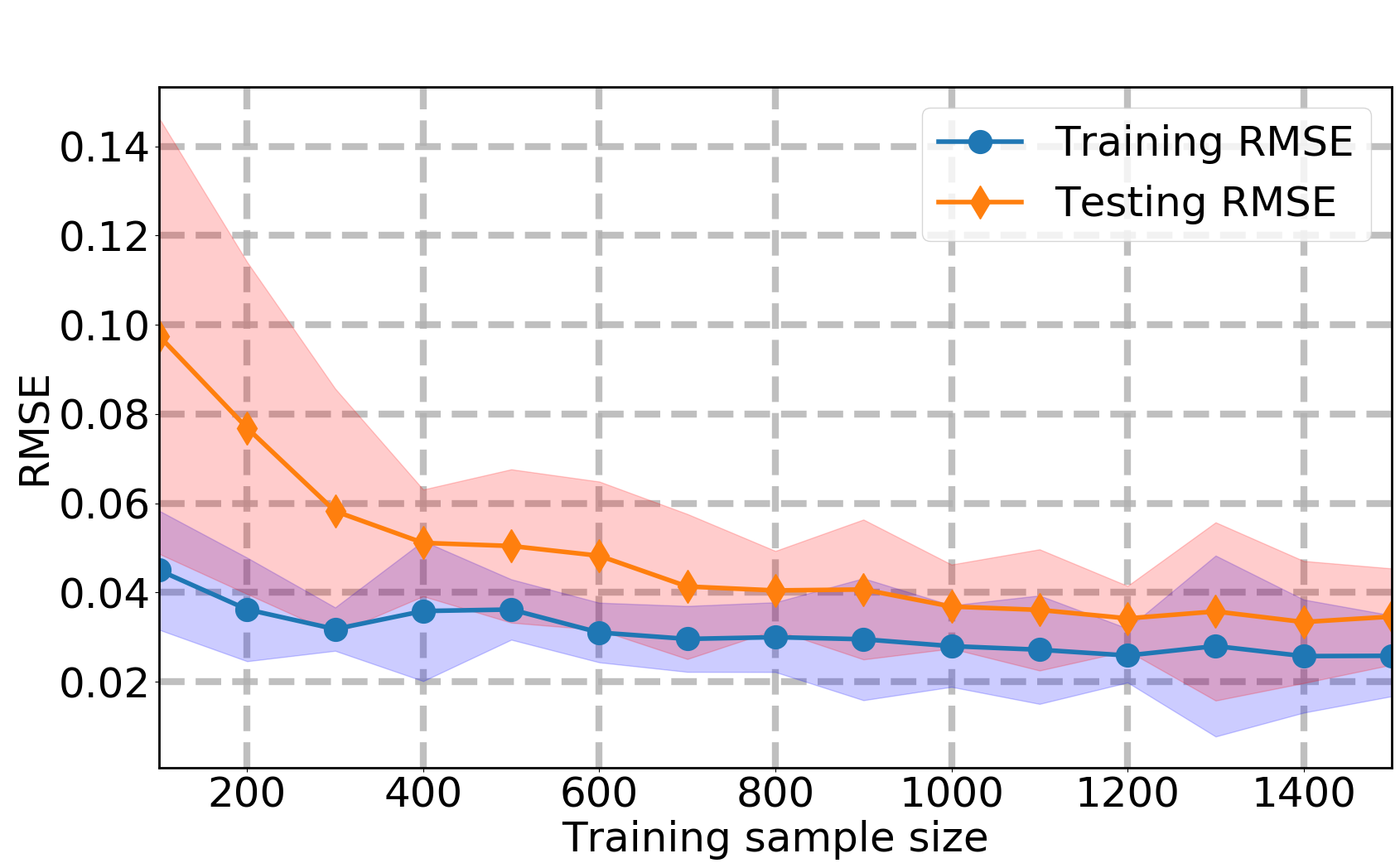}}
\subfigure[Pearson correlation]{
\centering
\includegraphics[width = 0.32\linewidth]{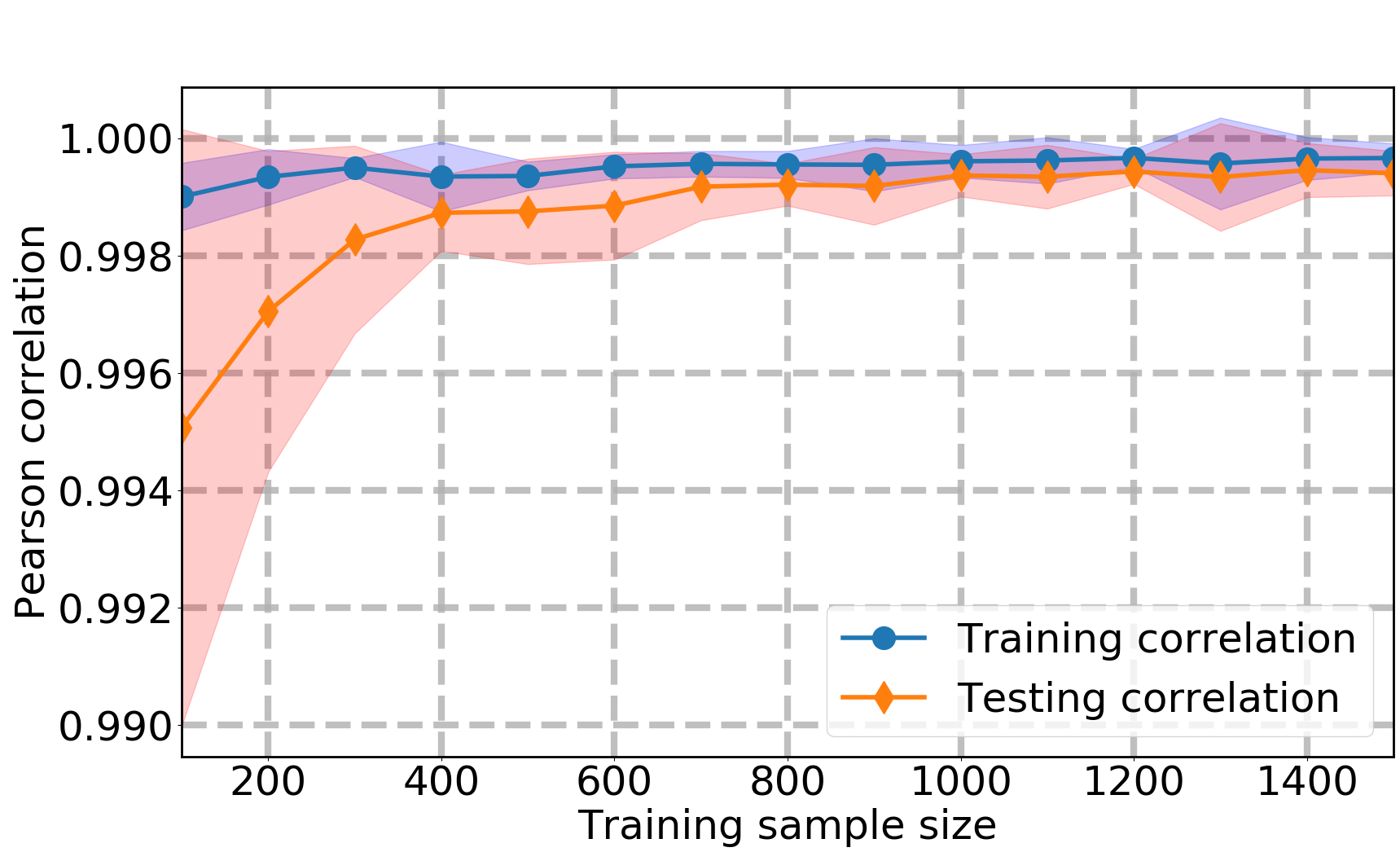}}
\subfigure[Crossplot]{
\centering
\includegraphics[width = 0.32\linewidth]{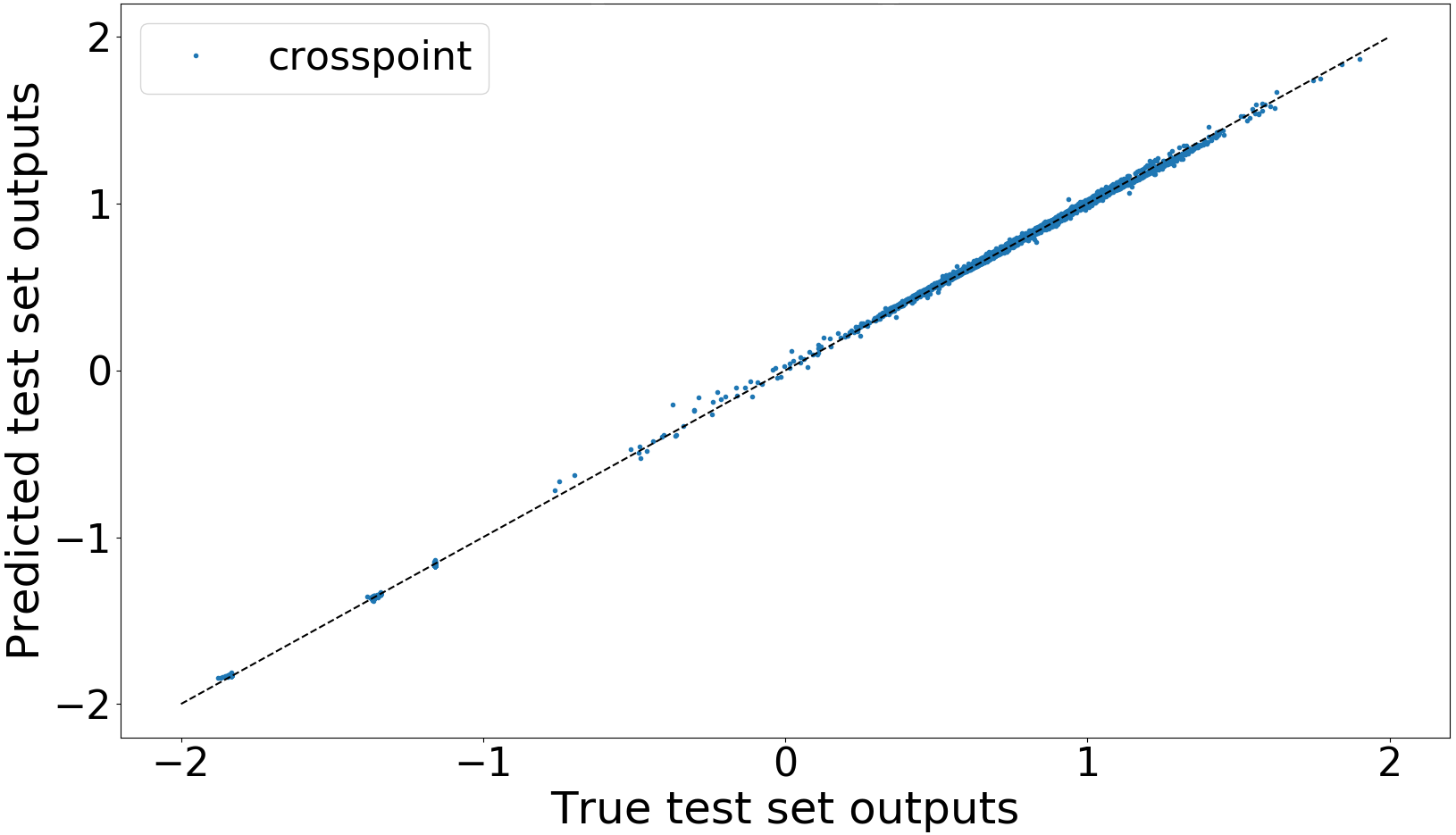}}
\caption{Sub-figure (a) and (b) shows RMSE/Correlation learning curve, respectively, in a function of training sample size. The blue/orange (darker/lighter in grey scale) curves denote performance in training/testing data set, respectively. The corresponding shadowed areas show its fluctuation in 2 standard deviations range. Sub-figure (c) shows the cross-plot between surrogate and simulator outputs.}
\label{fig:SPE01Learncurve}
\end{figure*}

\subsection{Industrial example: Black-Oil dynamic model}
The second test example employs a widely recognised and studied industrial Black-Oil reservoir simulator in geophysical applications. The simulator contains a number of partial differential equations to model fluid dynamics in a petroleum reservoir. Please see \citep{odeh1981comparison} and \citep{ramirez2017sampling, das2017fast} if one is interested in details of this simulator and its geophysical background. The tested Black-Oil simulator has 8 inputs in $\mtheta$ that describes subsurface geophysical properties and 60 outputs (6 features per time step for 10 time steps). Prior of $\mtheta$ follows uniform distribution, and its prior ranges and ground truths $\mtheta_{\ast}$ are listed in Table \ref{table:SPE01Truth}. The MNLHC scheme is adopted for data preparation. DRN with $\eta = 10$ is used for training with $100$ epochs $20$ mini-batch size. Other settings are kept the same as those in the previous examples, and the reported accuracy is an averaged over 10 folds CV results. Regarding MultiNest settings, the number of live samples ${\rm N_{live}}$ is set to $300$ with sampling efficiency ${\rm erf} = 0.8$ and tolerance ${\rm tol} = 0.5$.

\begin{table}
\caption{List of ground truth $\mtheta_{\ast}$ values. The last two columns indicate upper and lower limits of uniform distribution.}
\begin{tabular}{l|ccc}
\hline
\textbf{Parameter}   & \textbf{Truth} $\mtheta_{\ast}$ & \textbf{Lower limit} & \textbf{Upper limit}  \\ \hline
1 & 1.21 & 0.2 & 5.0    \\
2  & 0.3 & 0.2 & 5.0     \\
3 & 3.0 & 0.2 & 5.0      \\
4 & 0.26 & 0.1 & 1.0     \\
5 & 0.64 & 0.1 & 1.0     \\
6 & 1.0 & 0.75 & 1.25      \\
7 & 0.8 & 0.75 & 1.25     \\
8 & 1.2 & 0.75 & 1.25    \\
\hline
\end{tabular}
\label{table:SPE01Truth}
\end{table}

\subsubsection{DRN surrogate accuracy}

Figure \ref{fig:SPE01Learncurve} (a) shows the surrogate accuracy in RMSE with training sample size varying between $100$ and $1500$. Training RMSE stays at the level around $0.032$, while testing RMSE stays at around $0.045$. Figure \ref{fig:SPE01Learncurve} (b) shows the corresponding correlation performance, and both training and testing correlations achieve high accuracy ($> 99.8 \%$). A cross-plot of surrogate prediction versus simulator true output in out-of-sample test is shown in Figure \ref{fig:SPE01Learncurve} (c), the performance of which is consistent to the learning curves.

\subsubsection{Comparing with existing network structures}

\begin{figure*}[ht]
\subfigure[RMSE comparison]{
\centering
\includegraphics[width = 0.32\linewidth]{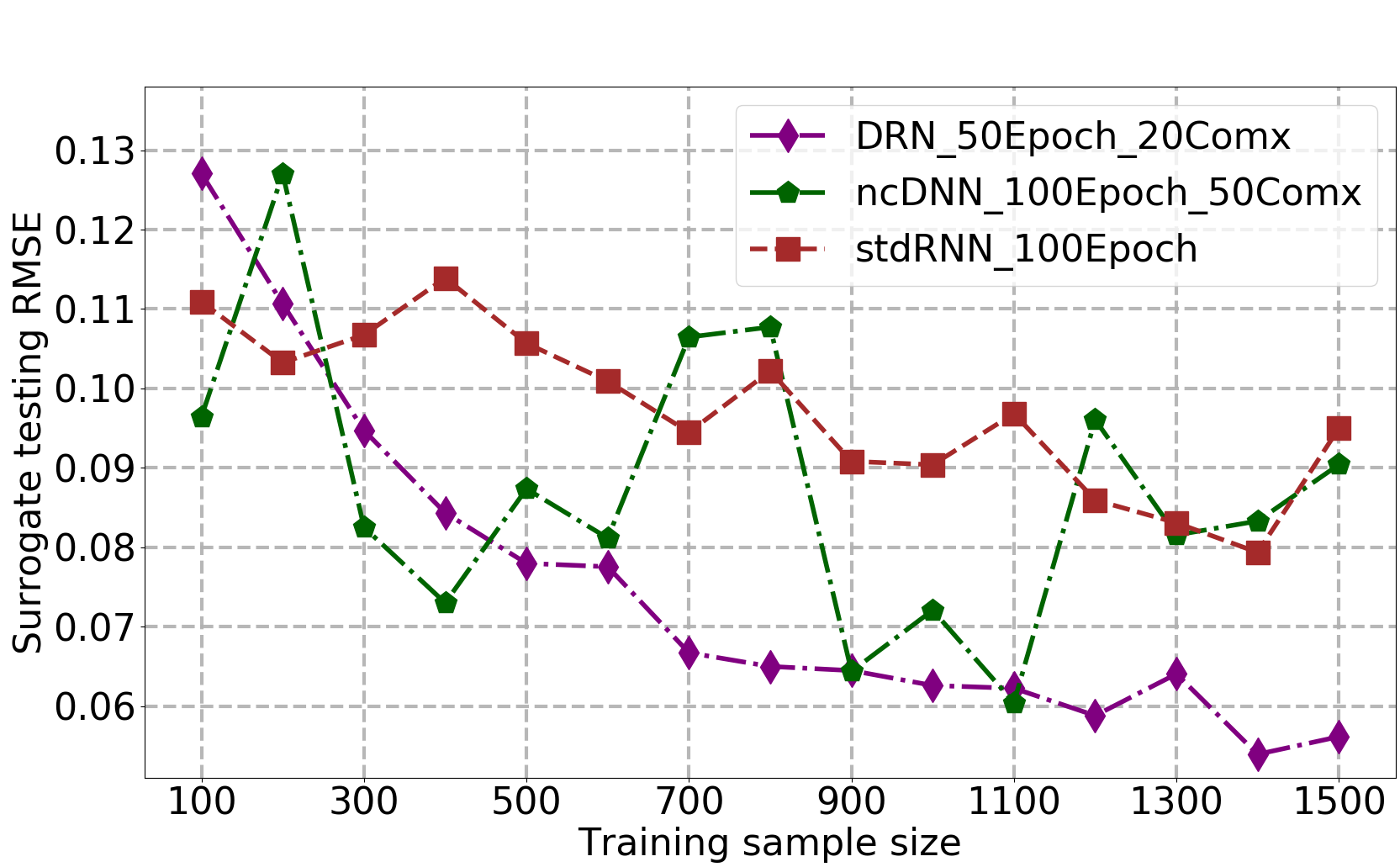}}
\subfigure[Model training time comparison]{
\includegraphics[width = 0.32\linewidth]{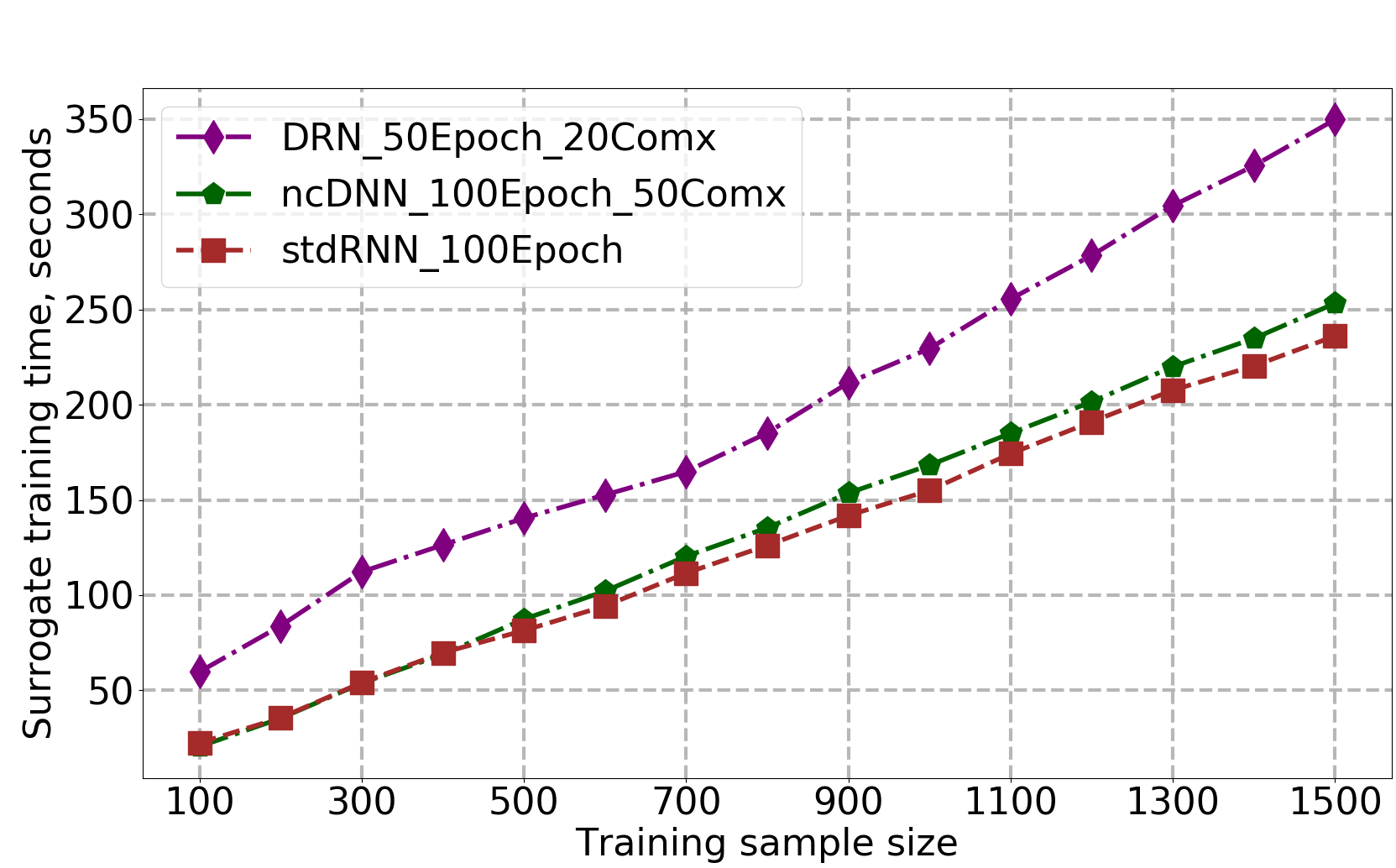}}
\subfigure[Total execution time comparison]{
\includegraphics[width = 0.32\linewidth]{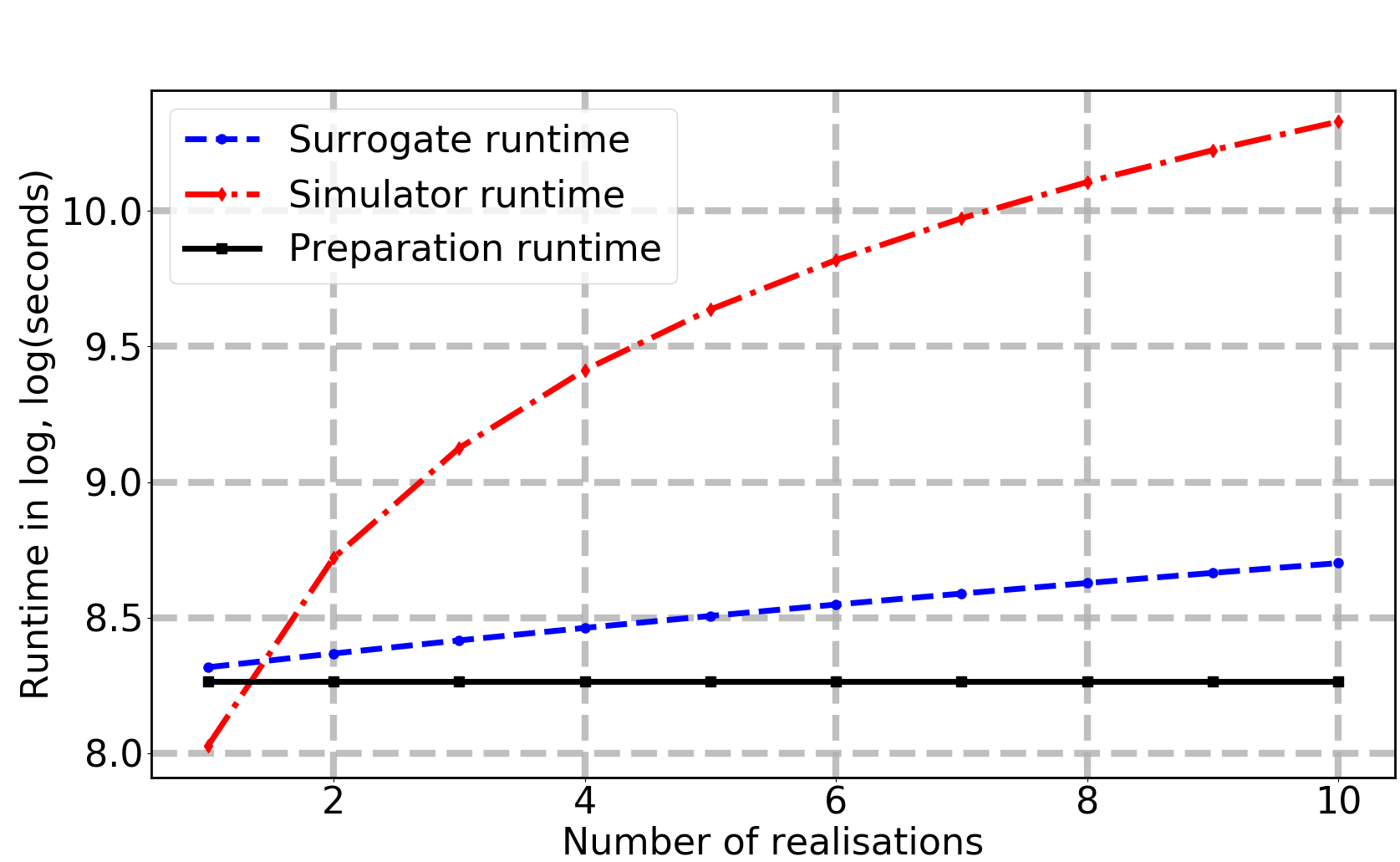}}
\caption{Sub-figure (a) and (b) shows comparison between different network structures in a function of training sample size. ${\rm DRN\_50Epoch\_20Comx}$' denotes DRN (purple curve/diamond marker) with $\eta = 20$ and 50 epochs. `ncDNN' denotes the non-cascading standard DNN structure, and `stdRNN' represents standard RNN with single hidden layer. Sub-figure (a) shows surrogate accuracy in RMSE, and sub-figure (b) shows model training time. Sub-figure (c) shows the total execution time comparison between the proposed surrogate and original simulation in a function of repeated realisations number (i.e., the number of repeats needed to re-run a task). The runtime is in log-scale, and the red dash-doted / blue dashed curves denote MultiNest estimation time with simulator/surrogate, respectively. The black flat line denotes a summation of data preparation time and surrogate training time per realisation.}
\label{fig:Exp2SurrCompare}
\end{figure*}

In addition to the proposed DRN, surrogate trained by other standard network structures are also implemented and evaluated. Specifically, a fully connected standard DNN with no cascading structure (ncDNN) and a standard RNN without extended hidden layers (stdRNN) are employed. ncDNN adopts a same DNN structure as the DNN component described previously, and its network complexity is set as $\eta = 50$. This non-cascading DNN takes $\mtheta$ as its input, and generate all 60 outputs simultaneously. stdRNN re-uses the recurrent structure described in Figure \ref{fig:compGraph}, but without extended DNN hidden component.

As shown in Figure \ref{fig:Exp2SurrCompare} (a), ncDNN performance is not robust for dynamic regression task, and the RMSE fluctuates with large variance. stdRNN shows a relatively stable RMSE performance, but its accuracy is much worse than that of DRN with 50 epochs. DRN's better performance is in a price of training time. As shown in Figure \ref{fig:Exp2SurrCompare} (b), DRN training time is clearly higher than the other two networks. Nevertheless, one sees that the training time increase rate is similar in all three methods, which suggests that DRN is still the best balanced choice for the the Black-Oil problem.  

\subsubsection{Estimation accuracy and overall runtime}

\begin{figure}[ht]
\centering
\includegraphics[width = 0.95\linewidth]{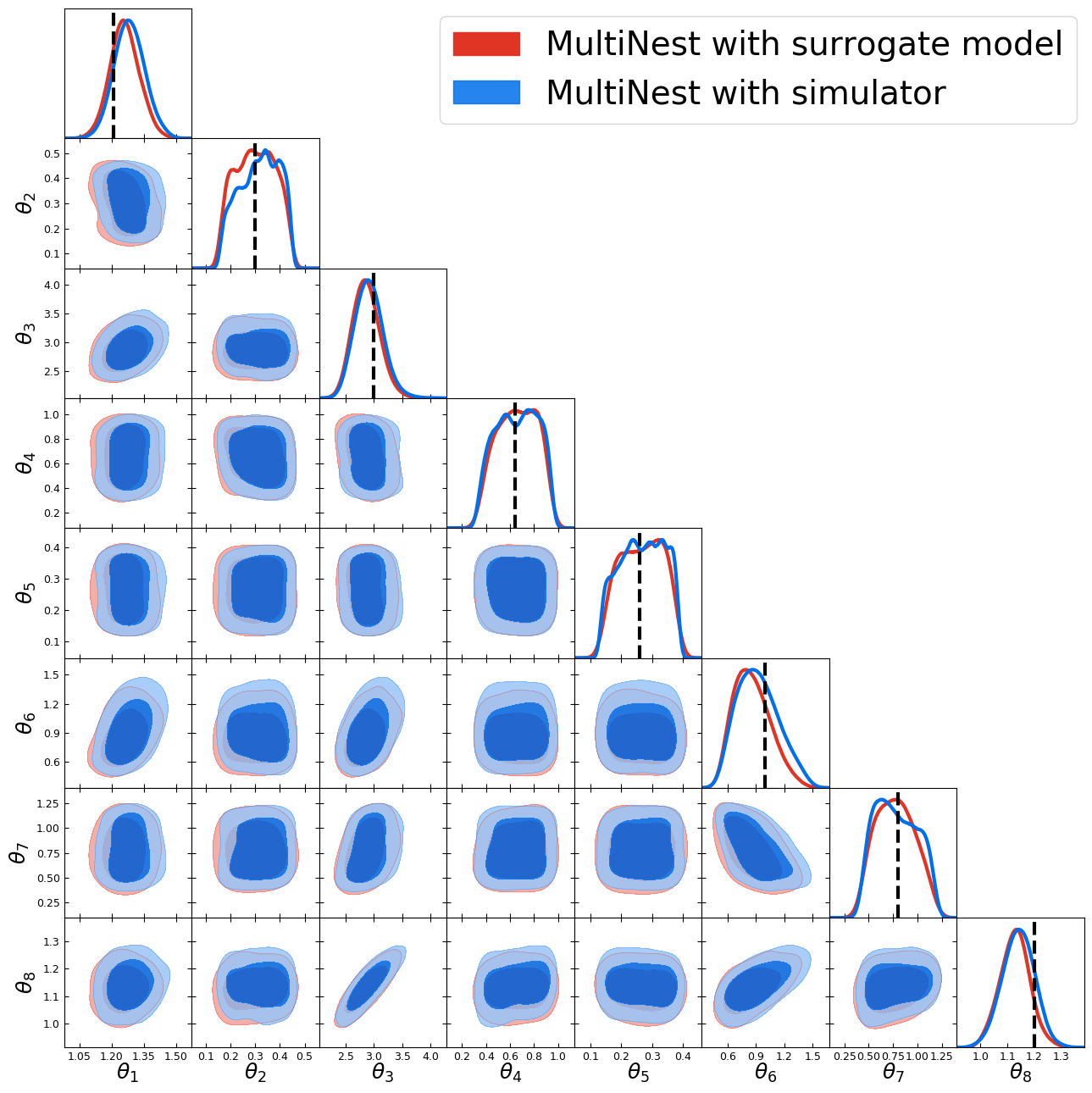}
\caption{Triangle plot of posterior estimation in $10\%$ noise. The diagonal blocks show posterior distributions of $\mtheta$. Curves and regions in red (darker) and blue (lighter) denote estimated posterior with DRN surrogate model and original simulator, respectively. The black dashed lines denote ground-truth of $\mtheta$.}
\label{fig:SPE01Triangle}
\end{figure}

Figure \ref{fig:SPE01Triangle} shows MultiNest estimation accuracy with $10\%$ noise in triangle plot. One sees that the peaks of posterior distributions in the triangle plot are very close to the ground-truth black dashed lines, and the coloured contour (red and blue areas) are almost overlapping with each other. All these results suggest the DRN trained surrogate model is highly consistent with the original simulator, and MultiNest can perform accurate posterior estimation through the trained surrogate model. 
Table \ref{table:Exp2_time} shows a comparison of approximated runtime between surrogate model and simulator. The training time is collected based on DRN structure with $\eta = 15$, $\rm epoch = 100$, and 1500 training data points. As shown in the second column of the table, the number of likelihood evaluations in both methods are comparable. The table clearly demonstrates that the proposed surrogate method can bring down MultiNest execution time for at least an order of magnitude.

\begin{table}
\caption{Runtime comparison between MultiNest with surrogate model versus original simulator. `Avg. numE': averaged likelihood evaluation number. `prepT': posterior data collection time. `trainT' and `execT' are surrogate training time and MultiNest execution time, respectively, and `sec.' denotes second.}
\label{table:Exp2_time}
\centering
\begin{tabular}{lc|ccc} \hline
Sampler   & Avg. numE & prepT  & trainT  & execT (sec.) \\ \hline
Surrogate & 12568 & 3062 & 819 & 213    \\  \hline
Simulator & 13345 & 0 & 0 & 3062 \\ \hline 
\end{tabular}
\end{table}

Figure \ref{fig:Exp2SurrCompare} (c) shows an approximation of total runtime in a function of repeated trail number. It demonstrates that the proposed method can save computational time (compare to the original simulator method) when the underlying example requires more than one repeated realisation. Note that this conclusion is based on specific settings and problems, the runtime can be affected by a number of factors including but not limit to: number of CPU/GPU cores, MultiNest hyper-parameter settings, and training sample size. 

\section{Conclusions and discussions} \label{Sec:Con}

This paper introduced a complete Bayesian surrogate learning methodology for fast parameter estimation in dynamic simulator-based regression problems. The proposed method has some limitations. Firstly, it is powerful for scenarios that require some, rather than one, simulator runs. Surrogate approach doesn't take advantage if only very few simulator runs are needed during a fairly long period of time (e.g. 3 times per year). Secondly, the proposed approach is not suitable for dynamic simulators that has large number of time steps, this is because each time step needs a separate DNN component training within the DRN. The proposed pipeline is flexible in changing its sampling and training algorithm components. For instance, the NS algorithm can be replaced by other sampling algorithms such as MCMC (For readers
interested in a comparison of MCMC and MultiNest, please see
\citep{chen2018improving} for more details), and the proposed DRN can also be replaced by other ML algorithms such as Gaussian Process, depending on specific needs. Since it is an emulator to a system simulator, there's no need to re-train a surrogate when new observation data becomes available.   

% \subsection{Software and Data}

% We strongly encourage the publication of software and data with the
% camera-ready version of the paper whenever appropriate. This can be
% done by including a URL in the camera-ready copy. However, do not
% include URLs that reveal your institution or identity in your
% submission for review. Instead, provide an anonymous URL or upload
% the material as ``Supplementary Material'' into the CMT reviewing
% system. Note that reviewers are not required to look at this material
% when writing their review.

% % Acknowledgements should only appear in the accepted version.
\section*{Acknowledgements}
The authors would like to thank Dr. Benjamin Ramirez, Dr. Adam Eales, and Mr.Paul Gelderblom for the helpful discussions on this topic.
% \textbf{Do not} include acknowledgements in the initial version of
% the paper submitted for blind review.

\clearpage
\bibliographystyle{icml2019}
\bibliography{bibFile}

\end{document}